\documentclass{article} 
\usepackage{nips15submit_e,times}
\usepackage{hyperref}
\usepackage{url}

\usepackage{amsmath,amsthm,amssymb,amsfonts}
\usepackage{graphicx, fullpage, color}
\usepackage{url}

\usepackage{epstopdf} 

\usepackage{mathtools}
\usepackage{multirow}
\usepackage{array}
\usepackage[colorinlistoftodos,bordercolor=orange,backgroundcolor=orange!20,linecolor=orange,textsize=scriptsize]{todonotes}
\usepackage{algpseudocode,algorithm}
\usepackage[shortlabels]{enumitem}
\usepackage{xspace}

\newcommand{\R}{\mathbb{R}}

\newcommand{\eqdef}{\overset{\text{def}}{=}}

\def\<#1,#2>{\left\langle #1, #2 \right\rangle}

\newcommand{\pp}{\mathcal{P}}
\newcommand{\FedOpt}{Federated Optimization\xspace}
\newcommand{\fedopt}{federated optimization\xspace}

\newcommand{\algname}{DSVRG\xspace}  
\newcommand{\tw}{\tilde{w}}
\newcommand{\node}{node\xspace} 
\newcommand{\nodes}{nodes\xspace}
\newcommand{\iid}{IID\xspace}

\title{Federated Optimization: \\ Distributed Optimization Beyond the Datacenter}

\author{
Jakub Kone\v{c}n\'{y}\thanks{Work performed while at Google, Inc.} \\
School of Mathematics \\
University of Edinburgh \\
\texttt{J.Konecny@sms.ed.ac.uk} \\
\And
H. Brendan McMahan \\
Google, Inc. \\
Seattle, WA 98103 \\
\texttt{mcmahan@google.com}
\And
Daniel Ramage \\
Google, Inc. \\
Seattle, WA 98103 \\
\texttt{dramage@google.com}}

\nipsfinalcopy 

\begin{document}

\maketitle

\begin{abstract}
We introduce a new and increasingly relevant setting for distributed optimization in machine learning, where the data defining the optimization are distributed (unevenly) over an extremely large number of \nodes, but the goal remains to train a high-quality centralized model. We refer to this setting as \emph{\FedOpt}. In this setting, communication efficiency is of utmost importance.

A motivating example for \fedopt arises when we keep the training data
locally on users' mobile devices rather than logging it to a data
center for training. Instead, the mobile devices are used as \nodes
performing computation on their local data in order to update a global
model. We suppose that we have an extremely large number of devices in
our network, each of which has only a tiny fraction of data available
totally; in particular, we expect the number of data points available
locally to be much smaller than the number of devices. Additionally,
since different users generate data with different patterns, we assume
that no device has a representative sample of the overall
distribution.

We show that existing algorithms are not suitable for this setting, and propose a new algorithm which shows encouraging experimental results. This work also sets a path for future research needed in the context of \fedopt.
\end{abstract}

\section{Introduction and Problem Formulation}
\label{sec:intro}

The optimization community has seen an explosion of interest in solving problems with finite-sum structure in recent years. In general, the objective is formulated as
\begin{equation}
\label{eq:problem}
\min_{w \in \R^d} f(w) \qquad \text{where} \qquad f(w) \eqdef \frac{1}{n} \sum_{i=1}^n f_i(w).
\end{equation}
The main source of motivation are problems arising in machine learning. The problem structure~\eqref{eq:problem} covers linear or logistic regressions, support vector machines, but also more complicated algorithms such as conditional random fields or neural networks. The functions $f_i$ are typically specified via a loss function $\ell$  dependent on a pair of input-output data point $\{x_i, y_i\}$,  $f_i(w) = \ell(w; x_i, y_i)$.

The amount of data that businesses, governments and academic projects collect is rapidly increasing. Consequently, solving the problem~\eqref{eq:problem} is becoming impossible with a single \node, as merely storing the whole dataset on a single \node becomes infeasible. This gives rise to the distributed optimization setting, where we solve the problem~\eqref{eq:problem} even though no single \node has direct access to all the data. 

In the distributed setting, communication cost is by far the largest bottleneck, as exhibited by a large amount of existing work. 
%
Further, many state-of-the-art optimization algorithms are inherently sequential, relying on a large number of very fast iterations. The problem stems from the fact that if one needs to perform a round of communication after each iteration, practical performance drops down dramatically, as the round of communication is much more time-consuming than a single iteration of the algorithm.

Works including \cite{balcanDistributedLearning, hydra, hydra2, shamir2014distributed} have established basic theoretical results and developed distributed versions of existing methods, but as this is a relatively new area, many open questions remain. Recently, the idea of communication efficient algorithms has gained traction. Such algorithms perform a large amount of computation locally, before each round of communication, ideally balancing the cost of computation and communication\cite{DANE, tianbao, cocoaNIPS, cocoaICML, DiSCO}. There has been an attempt at a thorough understanding of communication lower bounds \cite{CommShamir}, but the authors note that significant gaps remain in a number of settings.

\section{Federated Optimization --- The Challenge of Learning from Decentralized Data}
\label{sec:challenge}

With the exception of CoCoA \cite{cocoaNIPS, cocoaICML}, the existing work concerning communication efficient optimization \cite{DANE, DiSCO} and/or statistical estimation \cite{zhang2012communication, zhang2013information} presupposes, and in fact requires, that the number of \nodes is (much) smaller than the number of datapoints available to each \node, and that each \node has access to a random sample from the same distribution, and typically that each \node has an identical number of data points.

We are hoping to bring to attention a new and increasingly relevant setting for distributed optimization, where typically none of the above assumption are satisfied, and communication efficiency is of utmost importance. In particular, algorithms for \fedopt must handle training data with the following characteristics:
\begin{itemize}[noitemsep]
\item \textbf{Massively Distributed}: Data points are stored across a very large number of \nodes $K$. In particular, the number of \nodes can be much bigger than the average number of training examples stored on a given \node ($n/K$).
\item \textbf{Non-\iid}: Data on each \node may be drawn from a different distribution; that is, the data points available locally are far from being a representative sample of the overall distribution.
\item \textbf{Unbalanced}: Different \nodes may vary by many orders of magnitude in the number of training examples they hold.
\end{itemize}
In this work, we are particularly concerned with \textbf{sparse} data, where some features occur on only a small subset of \nodes or data points. We show that the sparsity structure can be used to develop an effective algorithm for \fedopt.

We are motivated by the setting where training data lives on users' mobile devices (phones and tablets), and the data may be privacy sensitive. The data $\{x_i, y_i\}_{i=1}^n$ are generated by device usage, e.g. interaction with apps. Examples include predicting the next word a user will type (language modeling for smarter keyboard apps), predicting which photos a user is most likely to share, or predicting which notifications are most important. 
To train such models using traditional distributed learning algorithms, one would collect the training examples in a centralized location (data center) where it could be shuffled and distributed evenly over compute nodes. Federated optimization provides an alternative model potentially saving significant network bandwidth and providing additional privacy protection. In exchange, users allow some use of their devices' computing power.

Communication constraints arise naturally in the massively distributed setting, as network connectivity may be limited (e.g., we may wish to defer all communication until the mobile device is charging and connected to a wi-fi network).  Thus, in realistic scenarios we may be limited to only a single round of communication per day. This implies that, within reasonable bounds, we have access to essentially unlimited local computational power. As a result, the practical objective is solely to minimize the rounds of communication.

Formally, let $K$ be the number of \nodes. Let $\{\pp_k\}_{k=1}^K$ denote a partition of data point indices $\{1, \dots, n\}$, so $\pp_k$ is the set of data points stored on \node $k$, and define $n_k = |\mathcal{P}_k|$. 
We can then rephrase the objective~\eqref{eq:problem} as a linear combination of the local empirical objectives $F_k(w)$, defined as 
\begin{equation*}
\label{eq:problem:distributed}
f(w) = \sum_{k = 1}^K \frac{n_k}{n} F_k(w) \eqdef \sum_{k=1}^K \frac{n_k}{n} \cdot \frac{1}{n_k} \sum_{i \in \mathcal{P}_k} f_i(w).
\end{equation*}

\section{An Algorithm for \FedOpt}

The main motivation for the algorithm we propose comes from a perhaps surprising connection between two algorithms --- SVRG \cite{SVRG, S2GD}
and DANE \cite{DANE}. SVRG is a stochastic method with variance reduction for solving the problem \eqref{eq:problem} on a single \node. DANE is an algorithm for distributed optimization which on every iteration solves exactly a new subproblem on each node based on the local data and the gradient of the entire function $f$.

One could modify the DANE algorithm in the following way: instead of solving the DANE subproblem exactly, use any optimization algorithm to produce an approximate solution. In particular, if we use SVRG as the local solver, the sequence of updates is \emph{equivalent} to the following version of distributed SVRG.

\begin{algorithm}[!h]
\begin{algorithmic}[1]
\State \textbf{parameters:} $m$ = \# of stochastic steps per epoch, $h$ = stepsize, data partition $\{\pp_k\}_{k=1}^K$, starting point $\tw_0$
\For {$s = 0, 1, 2, \dots$}
	\Comment Overall iterations
	\State Compute $\nabla f(\tw_s) = \frac{1}{n} \sum_{i=1}^n \nabla f_i(\tw_s)$
	\For {$k = 1$ to $K$} \textbf{in parallel} over \nodes $k$
	\Comment Distributed loop
	\State Initialize: $w_k = \tw_s$
	\For {$t = 1$ to $m$}
		\Comment Actual update loop
		\State Sample $i \in \mathcal{P}_k$ uniformly at random
		\State $ w_k \gets w_k - h \left( \nabla f_i(w_k) - \nabla f_i(\tw_s) + \nabla f(\tw_s) \right) $
		\EndFor
	\EndFor
	\State $\tilde{w} \gets \tilde{w} + \frac{1}{K} \sum_{k=1}^K (w_k - \tilde{w})$
	\Comment Aggregate updates from nodes
\EndFor
\end{algorithmic}

\caption{A distributed version of SVRG}
\label{alg:DS2GDnaive}
\end{algorithm}

The algorithm~\ref{alg:DS2GDnaive} indeed works very well in many ``simple'' settings, but fails in the challenging setting of \fedopt, particularly with sparse data. In order to make the algorithm more robust, we modify it in a number of ways. The most important changes include 
\begin{enumerate}[noitemsep]
\item Flexible stepsize $h_k$ --- different for each \node, inversely proportional to size of the local data, $n_k$
\item Scaling of stochastic updates by a diagonal matrix $S_k$ (defined below). Step 8 in Algorithm~\ref{alg:DS2GDnaive} is replaced by \hbox{$ w_k = w_k - h_k \left( S_k \left[ \nabla f_i(w_k) - \nabla f_i(\tilde{w}) \right] + \nabla f(\tilde{w}) \right) $}
\item The aggregation procedure is adaptive to the data distribution. For some diagonal matrix $A$ (defined below), the step 11 in Algorithm~\ref{alg:DS2GDnaive} is replaced by $\tilde{w} = \tilde{w} + A \sum_{k=1}^K \frac{n_k}{n} (w_k - \tilde{w})$
\end{enumerate}

The matrices $S_k$, and $A$ concern sparsity patterns in the data, and are identities in the case of fully dense data. 
To motivate their use, imagine that the data are distributed in the following way. All of the data points that have non-zero value of a certain feature are stored on a single \node $k$. Consequently, computing stochastic gradient $\nabla f_i(w_k)$ will greatly overestimate the gradient in this direction, because it appears much more frequently on \node $k$ than in the entire dataset. Diagonal elements of the matrix $S_k$ are ratios of frequencies of appearance of each feature globally and locally. One can interpret this as scaling the estimates so that they are of the correct magnitude in expectation, taken \emph{conditioned} on knowledge of the distribution of the sparsity patterns.

In order to motivate use of the matrix $A$, let us assume our problem is separable in the sense that each data point depends only on features belonging to one of a few disjoint clusters of coordinates, and data are distributed according to these clusters. In the case of linear predictors, we could simply solve the problems independently, and add up the results, instead of averaging, to obtain the optimal solution. Although this in not the case in reality, we try to interpolate, and average updates for features that appear on every \node, while we take longer steps for features that appear only on a few \nodes. Formally, $A_{ii} = K / \omega_i$, where $\omega_i$ is the number of \nodes on which feature $i$ is present. Although this is a heuristic modification, it's omission greatly degrades performance of the algorithm.

\section{Experiments}

The dataset presented here was generated based on public posts on a large social network. We randomly picked $10,000$ authors that have at least $100$ public posts in English, and try to predict whether a post will receive at least one comment (that is, a binary classification task). 

We split the data chronologically on a per-author basis, taking the earlier $75\%$ for training and the following $25\%$ for testing. The total number of training examples is $n = 2,166,693$. We created a simple bag-of-words language model, based on the $20,000$ most frequent words.
We then use a logistic regression model to make a prediction based on these features.

We shape the distributed optimization problem as follows. Suppose that each user corresponds to one \node, resulting in $K = 10,000$. The average $n_k$, number of data points on \node $k$, is thus approximately $216$. However, the actual numbers $n_k$ range from $75$ to $9,000$, showing the data distribution is in fact substantially unbalanced. It is natural to expect that different users can exhibit very different patterns in the data generated. This is indeed the case, and hence the distribution to \nodes cannot be considered an \iid sample from the overall distribution.

\begin{figure}[!h]
\centering
\includegraphics[width=0.49\textwidth]{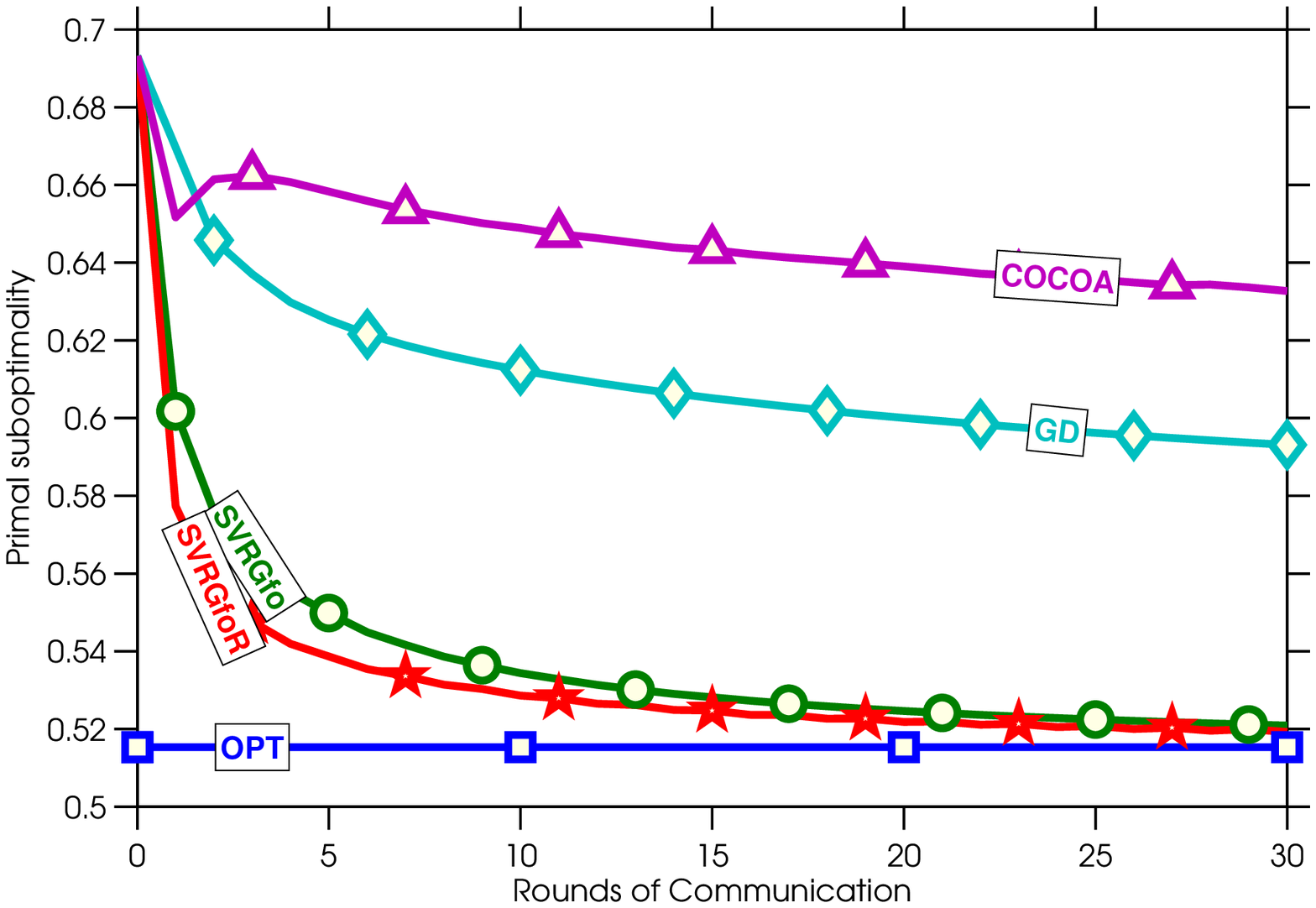}
\includegraphics[width=0.49\textwidth]{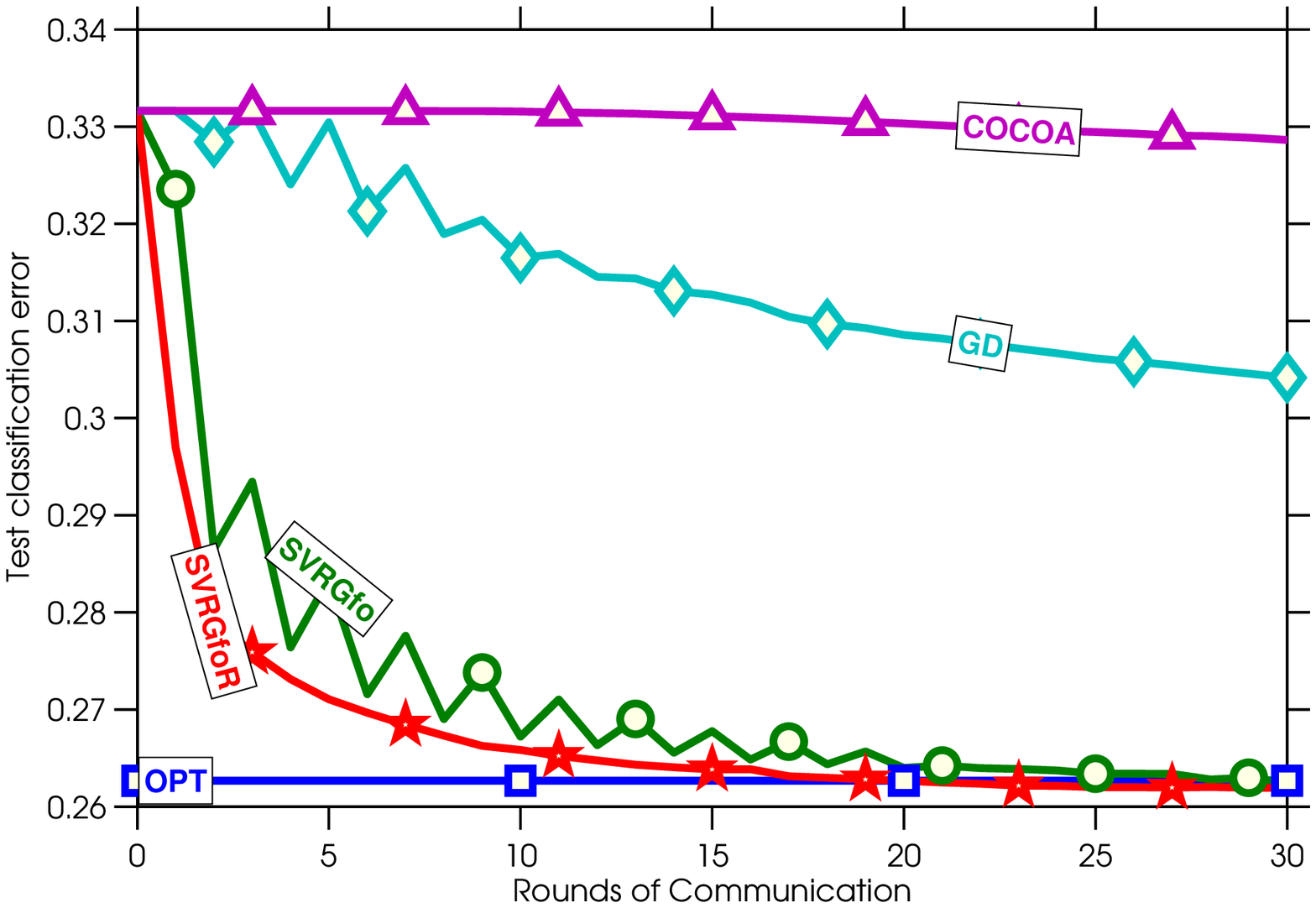}
\caption{Rounds of communication vs. objective function (left) and test prediction error (right).}
\label{fig:ex_final_001}
\end{figure}

In Figure~\ref{fig:ex_final_001}, we compare various optimization algorithms. Two other communication efficient algorithms, DANE \cite{DANE} and DiSCO\cite{DiSCO}, diverge in this setting, and so are not included in the figure.
\begin{itemize}[noitemsep]
 \item The blue squares (OPT) represent the best possible offline value (the optimal value of the optimization task in the first plot, and the test error corresponding to the optimum in the second plot).
 \item  The teal diamonds (GD) correspond to a simple distributed gradient descent. 
 \item The purple triangles (COCOA)  are for the CoCoA algorithm \cite{cocoaNIPS}. 
\item The green circles (SVRGfo) give values for our proposed algorithm.
\item The red stars (SVRGfoR) correspond to the same algorithm applied to the same problem with randomly reshuffled data.  That is, we keep the unbalanced number of examples per node, but populate each node with randomly selected examples.
\end{itemize}

The experiments point at several important realities. First, existing communication efficient algorithms are completely inefficient for \fedopt. The only algorithm that converges, CoCoA, does so at a slower rate than a naive benchmark --- distributed gradient descent. Our algorithm, \algname, reaches optimality in very few communication rounds, despite the challenging setting of \fedopt. Furthermore, very little performance is lost compared to the setting where the data are randomly reshuffled between \nodes, which highlights the robustness of our proposed method to non-IID data.

We argue that the setting of \fedopt will be increasingly important for practical application, as mobile devices get computationally more powerful and privacy issues ever more pressing. A large number of question remain open, creating potential for a line of further work. Most importantly, there is no readily available dataset (from usual sources as libsvm or UCI repository) with naturally user-clustered data. Creation of a new, public dataset is crucial to lower the barrier of further engagement of the academic community. More rigorous experiments remain to be done, both on larger datasets and on more challenging problems, such as deep learning. Currently, there is no proper theoretical justification of the method, which would surely drive further improvements. Lastly, in order for this shift to ``on-device intelligence'' to truly happen, methods from differential privacy need to be incorporated and many parts of practical machine learning pipelines need to be redesigned.

\bibliography{notes}

\begin{thebibliography}{10}

\bibitem{CommShamir}
Yossi Arjevani and Ohad Shamir.
\newblock Communication complexity of distributed convex learning and
  optimization.
\newblock {\em arXiv preprint arXiv:1506.01900}, 2015.

\bibitem{balcanDistributedLearning}
Maria-Florina Balcan, Avrim Blum, Shai Fine, and Yishay Mansour.
\newblock Distributed learning, communication complexity and privacy.
\newblock {\em arXiv preprint arXiv:1204.3514}, 2012.

\bibitem{hydra2}
Olivier Fercoq, Zheng Qu, Peter Richt{\'a}rik, and Martin Tak{\'a}c.
\newblock Fast distributed coordinate descent for non-strongly convex losses.
\newblock In {\em Machine Learning for Signal Processing (MLSP), 2014 IEEE
  International Workshop on}, pages 1--6. IEEE, 2014.

\bibitem{cocoaNIPS}
Martin Jaggi, Virginia Smith, Martin Tak{\'a}c, Jonathan Terhorst, Sanjay
  Krishnan, Thomas Hofmann, and Michael~I Jordan.
\newblock Communication-efficient distributed dual coordinate ascent.
\newblock In {\em Advances in Neural Information Processing Systems}, pages
  3068--3076, 2014.

\bibitem{SVRG}
Rie Johnson and Tong Zhang.
\newblock Accelerating stochastic gradient descent using predictive variance
  reduction.
\newblock In {\em Advances in Neural Information Processing Systems}, pages
  315--323, 2013.

\bibitem{S2GD}
Jakub Kone{\v{c}}n{\'y} and Peter Richt{\'a}rik.
\newblock Semi-stochastic gradient descent methods.
\newblock {\em arXiv preprint arXiv:1312.1666}, 2013.

\bibitem{cocoaICML}
Chenxin Ma, Virginia Smith, Martin Jaggi, Michael~I Jordan, Peter
  Richt{\'a}rik, and Martin Tak{\'a}{\v{c}}.
\newblock Adding vs. averaging in distributed primal-dual optimization.
\newblock {\em arXiv preprint arXiv:1502.03508}, 2015.

\bibitem{hydra}
Peter Richt{\'a}rik and Martin Tak{\'a}{\v{c}}.
\newblock Distributed coordinate descent method for learning with big data.
\newblock {\em arXiv preprint arXiv:1310.2059}, 2013.

\bibitem{shamir2014distributed}
Ohad Shamir and Nathan Srebro.
\newblock Distributed stochastic optimization and learning.
\newblock In {\em Communication, Control, and Computing (Allerton), 2014 52nd
  Annual Allerton Conference on}, pages 850--857. IEEE, 2014.

\bibitem{DANE}
Ohad Shamir, Nathan Srebro, and Tong Zhang.
\newblock Communication efficient distributed optimization using an approximate
  newton-type method.
\newblock {\em arXiv preprint arXiv:1312.7853}, 2013.

\bibitem{tianbao}
Tianbao Yang.
\newblock Trading computation for communication: Distributed stochastic dual
  coordinate ascent.
\newblock In {\em Advances in Neural Information Processing Systems}, pages
  629--637, 2013.

\bibitem{zhang2013information}
Yuchen Zhang, John Duchi, Michael~I Jordan, and Martin~J Wainwright.
\newblock Information-theoretic lower bounds for distributed statistical
  estimation with communication constraints.
\newblock In {\em Advances in Neural Information Processing Systems}, pages
  2328--2336, 2013.

\bibitem{zhang2012communication}
Yuchen Zhang, Martin~J Wainwright, and John~C Duchi.
\newblock Communication-efficient algorithms for statistical optimization.
\newblock In {\em Advances in Neural Information Processing Systems}, pages
  1502--1510, 2012.

\bibitem{DiSCO}
Yuchen Zhang and Lin Xiao.
\newblock Communication-efficient distributed optimization of self-concordant
  empirical loss.
\newblock {\em arXiv preprint arXiv:1501.00263}, 2015.

\end{thebibliography}

\end{document}